\documentclass[10pt,twocolumn,letterpaper]{article}

\usepackage{cvpr}
\usepackage{times}
\usepackage{epsfig}
\usepackage{graphicx}
\usepackage{amsmath}
\usepackage{amssymb}
\usepackage{multirow}
\usepackage{footnote}

\usepackage[pagebackref=true,breaklinks=true,letterpaper=true,colorlinks,bookmarks=false]{hyperref}

\cvprfinalcopy 

\def\cvprPaperID{****} 

\ifcvprfinal\fi
\begin{document}

\title{ Deep Local Video Feature for Action Recognition}

\author{
Zhenzhong Lan\\
Carnegie Mellon University\\
{\tt\small lanzhzh@cs.cmu.edu}
\and
Yi Zhu\\
University of California, Merced \\
{\tt\small yzhu25@ucmerced.edu}
\and 
Alexander G. Hauptmann \\
Carnegie Mellon University\\
{\tt\small alex@cs.cmu.edu}
}

\maketitle


\begin{abstract}

We investigate the problem of representing an entire video using CNN features for human action recognition. Currently, limited by GPU memory, we have not been able to feed a whole video into CNN/RNNs for end-to-end learning. A common practice is to use sampled frames as inputs and video labels as supervision. One major problem of this popular approach is that the local samples may not contain the information indicated by global labels. To deal with this problem, we propose to treat the deep networks trained on local inputs as local feature extractors. After extracting local features, we aggregate them into global features and train another mapping function on the same training data to map the global features into global labels. We study a set of problems regarding this new type of local features such as how to aggregate them into global features. Experimental results on HMDB51 and UCF101 datasets show that, for these new local features,  a simple maximum pooling on the sparsely sampled features lead to significant performance improvement.

\end{abstract}

\section{Introduction}

Given the success of deep neural networks on image classification, we have been hoping that it can achieve similar improvement on video classification. However, after several years' effort, the hope remains elusive. The major obstacles, we believe, lie in two major differences between the images and videos. Compared to images, videos are often much larger in size and video labels are often much more expensive to get. Large data size means that it is difficult to feed a whole video into modern deep CNN/RNN architectures that often have large memory demands. Expensive video labeling brings difficulties in getting enough labeled data to train a robust network.  Recent approaches \cite{simonyan2014two,wanggoodpractice2015,TSN2016} circumvent these problems by learning on sampled frames or very short video clips (local inputs) with video level (global) labels. 

However, video level label information can be incomplete or even missing at frame/clip level. This information mismatch leads to the problem of false label assignment.  In other words, the imprecise frame/clip-level labels populated from video labels are too noisy to guide precise mapping from videos to labels. To deal with this problem, a common practice is to sample multiple frames/clips from a video at testing time and aggregate the prediction scores of these sampled frames/clips to get the final prediction for that video. However, simply averaging the prediction scores, without another level of mapping, is not enough to recover the damages brought by false label assignment.

To further correct the damages made by false label assignment, we propose to treat the deep networks that trained on local inputs as feature extractors. After extracting local features using these pre-trained networks, we aggregate them into global features and train another mapping function (shallow network) on the same training data to map the global features into global labels. 

Our method is very similar to the fine-tuning practices that have been popular in image classification. The major difference is that the data we used to train our feature extraction networks are local and the label are noisy due to the false label assignment. Therefore, we heavily rely on the shallow network to correct the mistakes we made on local feature learning. 

Our method is also similar to the practices of using ImageNet pre-trained networks to extract frame-level (local) features for video classification \cite{xu2015discriminative, lan2013cmu}. The only difference here is that our local feature extractors (deep networks) are trained on the same training data as the classifiers (shallow networks). However, this simple fine-tuning has great impact on the local features. First, fine-tuning on the video data narrows the domain-gap between feature extractors and the target data. Second, since we use the same training data twice,  the local features we extracted can be severely overfitted to the training data. 

A common practice to deal with the overfitting problem is to use cross-validation. However, in this particular case, we cannot use this method. This is largely because that we have already had the problem of lacking training data and cannot afford to lose any more training data. It is also because of the difficulties in calibrating features generated from different models. Therefore, we choose to use the whole training set to train the local feature extractors in the hope that those layers that are far away from the probability layers would capture general video information hence generalizable enough for further classifiers training.

We call this new type of local video features as Deep lOcal Video Features (DOVF). 

In summary, DOVF is a kind of local video features that are extracted from deep neural networks trained on local video clips and global video labels.  The major problems we would like to investigate about DOVF are:
\begin{itemize}
    \item Which layer(s) of features should we extract? Without further investigation, the only thing we know is that we cannot utilize the probability layer as it severely overfits to the noisy labels and would have a large distribution difference between features of training and testing sets. 
    \item How to aggregate the local features into global features? We will test various feature aggregation methods such as mean pooling and Fisher Vectors (FV).
    \item How dense should we extract the local features? In practice, we would prefer sparse temporal sampling as it would be much more efficient.
    \item How complementary are DOVF to the traditional local features such as IDT\cite{wang2013action}? The more complementary they are, the more room we can improve by incorporating methodologies that we developed for traditional local features.  
\end{itemize}

In the remainder of this paper, we first provide more background information about video features with an emphasis on recent attempts on learning with deep neural networks.  We then describe our experimental settings in detail. After that, we evaluate our methods on HMDB51 and UCF101 datasets. Further discussions including potential improvements are given at the end.

\section{Related works}

New video representation methods are the major sources of breakthroughs for video classification.

In traditional video representations, trajectory based approaches \cite{wang2013action,jiang2012trajectory}, especially the Dense Trajectory (DT) and IDT \cite{wang2011action, wang2013action}, are the basis of current state-of-the-art hand-crafted algorithms. These trajectory-based methods are designed to address the flaws of image-extended video features. Their superior performance validates the need for a unique representation of motion features. There have been many studies attempting to improve IDT due to its popularity. Peng et al. \cite{peng2014bag} enhanced the performance of IDT by increasing codebook sizes and fusing multiple coding methods. Sapienza et al. \cite{sapienza2014feature} explored ways to sub-sample and generate vocabularies for DT features. Hoai \& Zisserman \cite{hoai2014improving} achieved superior performance on several action recognition datasets by using three techniques including data augmentation, modeling score distribution over video subsequences, and capturing the relationship among action classes. Fernando et al. \cite{fernandomodeling} modeled the evolution of appearance in the video and achieved state-of-the-art results on the Hollywood2 dataset. \cite{lan2014beyond} proposed to extract features from videos with multiple playback speeds to achieve speed invariances. However, with the arising of deep neural network methods, these traditional methods are gradually forgotten. 

Motivated by this success of CNNs, researchers are working intensely towards developing CNN equivalents for learning video features. Several accomplishments have been reported from using CNNs for action recognition in videos \cite{zha2015exploiting, wu2015modeling, varadarajan2015efficient}. Karpathy et al. \cite{karpathy2014large} trained deep CNNs through one million weakly labelled YouTube videos and reported moderate success while using it as a feature extractor. Simonyan \& Zisserman \cite{simonyan2014two} demonstrated a result competitive to IDT \cite{wang2013action} through training deep CNNs using both sampled frames and stacked optical flows. Wang et al. \cite{wang2015action, wanggoodpractice2015, TSN2016} show multiple insightful analysis about how to improve two-stream frameworks and find several useful observations including pre-training two-stream ConvNets, using smaller learning rate, and using deeper networks, etc.  With these observations, they finally outperforms IDT \cite{wang2013action} by a large margin on UCF101 dataset. However, all these approaches rely on shot-clip predictions to get the final video scores without using global features. 

At the time we wrote this paper, two similar work \cite{diba2016deep, qiu2016deep} have been published on Arxiv. Both of them propose a new feature aggregation method to pool together the local neural network features into global video features. Diba et al. \cite{diba2016deep} proposes a bilinear model to pool together the outputs of last convolutional layers of the pre-trained networks and achieve state-of-the-art results on both HMDB51 and UCF101 datasets. Qiu et al. \cite{qiu2016deep} proposes a new quantization method that is similar to FV and achieves similar performance as \cite{diba2016deep}. However, neither of them provide detailed analysis of the local neural network features they have used. In this paper, we perform a more extensive analysis and show that a simple max pooling can achieve similar or better results compared to those much more complex feature aggregation methods as in \cite{diba2016deep, qiu2016deep}.

\section{Experimental settings}

\begin{table*}[!htbp]
\centering
\begin{tabular}{ |c|| c  c  c |c  c  c|}
 \hline
 \multirow{2}{4em}{ID} & \multicolumn{3}{|c|}{VGG16} & \multicolumn{3}{|c|}{Inception-BN}\\
  & Name & Dimensions & Type & Name & Dimension & Type  \\
 \hline
L-1 & fc8 & 101 & FC &fc-action & 101 & FC \\  
L-2 & fc7 & 4096 & FC & global\_pool  & 1024 & Conv\\
L-3 & fc6 & 4096 & FC & inception\_5b  & 50176 & Conv\\
L-4 & pool5 & 25088 & Conv & inception\_5a & 50176 & Conv\\
L-5 & conv5\_3  & 100352 & Conv & inception\_4e & 51744 & Conv\\
 \hline
\end{tabular}
\caption{Layers name and dimensions of those layers that we exam}
\label{table:layers_name}
\end{table*}

\begin{table*}[!htbp]
\centering
\begin{tabular}{ |c|| c c |c c |c c |}
 \hline
 \multirow{2}{4em}{Layers } & \multicolumn{2}{|c|}{Spatial Convnets ($\%$) } & \multicolumn{2}{|c|}{Temporal Convnets ($\%$) } & \multicolumn{2}{|c|}{Two-stream ($\%$)}  \\
  & VGG-16 & Inception-BN & VGG16 & Inception-BN & VGG-16 & Inception-BN\\
 \hline
L-1  & 77.8& 83.9 & 82.6& 83.7 & 89.6& 91.7  \\
L-2  & 79.5& 88.3 & 85.1& 88.8 & 91.4& 94.2 \\
L-3  & 80.1 & 88.3 & 86.6 & 88.7 & 91.8& 93.9 \\
L-4  & 83.7 & 85.6 & 86.5& 85.3 & 92.4& 91.4\\
L-5  & 83.5& 83.6 & 87.0& 83.6 & 92.3& 89.8\\
TSN \cite{TSN2016} & 79.8& 85.7 & 85.7& 87.9 & 90.9& 93.5 \\
 \hline
\end{tabular}
\caption{Layer-wise comparison of VGG-16 and Inception-BN networks on the split1 of UCF101 \label{layers}}
\end{table*}

For local feature extraction, we utilize both VGG16 and Inception-BN that were trained by Wang et al. \cite{wanggoodpractice2015, TSN2016}. We extract the outputs of the last five layers as our features from both networks. Table \ref{table:layers_name} shows the layer names of both networks and the correspondent feature dimensions. We classify these layers into two classes: full-connected (FC) layers and convolution (Conv) layers (pooling layers are treated as Conv layer). FC layers have much more parameters hence are much easier to overfit to the noisy labels than Conv layers. As shown, VGG16 has three FC layers while Inception-BN only has one.

Following the testing scheme of \cite{simonyan2014two,TSN2016}, we evenly sample 25 frames and flow clips for each video. For each frame/clip, we do 10x data augmentation by cropping the 4 corners and 1 center and performing their horizontal flipping from the cropped frames. After getting the features from the augmented data of each frame/clip, we average the features to get the local feature for that frame/clip. In the end, for each video, we get a total of 25 local features. The dimensions of each local features are shown in Table \ref{table:layers_name}. 

We use several local feature aggregation methods ranging from simple mean, maximum pooling to more complex feature encoding methods such as Bag of words ($BoW$), Vector of Locally Aggregated Descriptors ($VLAD$) and Fisher Vector ($FV$) encoding. To incorporate global temporal information, we divide each video into three parts and perform the aggregation for each part separately. For example, to aggregate the 25 local features of one video, we aggregate the first 8, middle 9, and the last 8 local features separately and concatenate the aggregated features to get the global feature. 

To map global features into global labels, we use SVM with Chi2 kernel and a fixed C=100 as in \cite{lan2014beyond} except for FV and VLAD, where we use linear kernel as suggested in \cite{vedaldi2012efficient}. To get the two-stream results, we fuse the prediction scores of spatial-net and temporal-net with the weights of 1 and 1.5 respectively, as in \cite{TSN2016}.

\section{Evaluation}

In this section, we experimentally answer the questions we raised in the introduction section using the results of both UCF101 and HMDB51 datasets. By default, we will use the outputs of global$\_$pool layer from Inception-BN network as our features and use maximum pooling to aggregate these local features into global features.  

\begin{table*}[!htbp]
\centering
\begin{tabular}{ |c|| c c |c c |c c |}
 \hline
 \multirow{2}{4em}{Layers }& \multicolumn{2}{|c|}{Spatial Convnets ($\%$) } & \multicolumn{2}{|c|}{Temporal Convnets ($\%$) } & \multicolumn{2}{|c|}{Two-stream ($\%$)}  \\
  & HMDB51 & UCF101 & HMDB51 & UCF101 & HMDB51 & UCF101\\
 \hline
$Mean$  & 56.0 & 87.5 & 63.7 & 88.3& 71.1 & 93.8    \\
$Mean\_Std$ & 58.1 & 88.1& 65.2 & 88.5& 72.0 & 94.2 \\
$Max$ &  57.7 & 88.3 & 64.8 & 88.8& 72.5 & 94.2 \\
$BoW$ &36.9 &71.9 & 47.9& 80.0& 53.4 & 85.3\\
$FV$ & 39.1 &69.8 & 55.6 & 81.3& 58.5 & 83.8\\
$VLAD$ & 45.3 &77.3 & 57.4 & 84.7& 64.7 & 89.2\\
 \hline
\end{tabular}
\caption{Comparison of different local feature aggregation methods on the split1 of UCF101 and HMDB51 \label{aggregate}}
\end{table*}

\begin{table*}[!htbp]
\centering
\begin{tabular}{ |c|| c c |c c |c c |}
 \hline
 \multirow{2}{4em}{\# of samples } & \multicolumn{2}{|c|}{Spatial Convnets ($\%$) } & \multicolumn{2}{|c|}{Temporal Convnets ($\%$) } & \multicolumn{2}{|c|}{Two-stream ($\%$)}  \\
  & HMDB51 & UCF101 & HMDB51 & UCF101 & HMDB51 & UCF101\\
 \hline
3  & 52.5 & 85.6 & 54.9 & 82.4& 64.6 & 91.6    \\
9 & 56.1 & 87.4& 62.2 & 87.7& 70.9 & 93.5 \\
15 &  56.9 & 88.2 & 64.4 & 88.5& 72.3& 93.8 \\
21 & 57.1 &88.1 & 64.8& 88.6& 71.8 & 94.1\\
25 &  57.7 & 88.3 & 64.8 & 88.8& 72.5 & 94.2 \\
Max & 57.6 & 88.4 & 65.3& 88.9& 72.4& 94.3 \\
 \hline
\end{tabular}
\caption{Number of samples versus accuracies\label{samples}}
\end{table*}

\subsection{Which layer(s) of features should we extract?}

To find out which (type of) layers we need to extract, we conduct experiments on both VGG16 and Inception-BN and show the results of split 1 of UCF101 in Table \ref{layers}.

 Table \ref{layers} shows that the the L-2 layer for Inception-BN and L-4 layer for VGG16 give the best performance. One common characteristic of these two layers is that they are the last convolution layers in both networks. There are three potential reasons for the superior performance of the last convolution layers. First, compared to the fully connected layers, the convolution layers have much less parameters, hence are much less likely to overfit to the training data that has false label assignment problem. Second, the fully-connected layers do not preserve the spatial information while those convolution layers do. Third, compared to other convolution layers that are further away from the probability layers, these layers contain more global information. Therefore, in terms of which layer(s) to extract, we suggest to extract the last convolution layer and avoid those full-connected layers. These results may justify why these recent works \cite{xu2015discriminative, diba2016deep, qiu2016deep} choose the outputs of the last convolution layers of the networks for further processing.

Compared to the results of Wang et al. \cite{TSN2016}, from which we get the pre-trained networks, we can see that our approach do improve the performance on both spatial-net and temporal-net. However, the improvements from spatial networks are much larger. This larger improvement may be because that, in training local feature extractors, the inputs for spatial net are single frames while the input for temporal net are video clips with 10 stacked frames. Smaller inputs lead to larger chance of false label assignment hence larger performance gap compared to our global feature approach. 

Previous works \cite{zha2015exploiting, lan2013cmu,xu2015discriminative} on using local features from ImageNet pre-trained networks show that combining features from multiple layers help to improve the overall performance significantly. We perform a similar analysis but found no improvement. This difference shows that fine-tuning brings some new characteristics to the local features. 

In the following experiments, we will only use the output of the global\_pool layer from inception-BN network, which has better performance than other layers in both networks.

\subsection{How to aggregate the local features into global features?}

To determine which aggregation methods is better, we test six aggregation methods on the split1 of both UCF101 and HMDB51 datasets.   

Assuming that we have $n$ local features, each of which has a dimension of $d$, the six different local feature aggregation methods are summarized as follows:

\begin{itemize}
    \item $Mean$, takes mean of these $n$ local features and produces a $1 \times d$ dimensional global feature. 
    \item $Max$, takes maximum of these $n$ local features along each dimension. 
    \item $Mean\_Std$, inspired by the Fisher Vector encoding, we records both mean and standard deviation of each dimension of the $n$ local features.
    \item $BoW$,  models the distribution of local features using $k$-means and quantizes them into these $k$ centroids.
    \item $VLAD$, models the distributions of local features using $k$-means and measures the mean of the differences between each local feature and the $k$ centroids.
    \item $FV$, models the distribution using GMMs with $k$ Gaussian and measures the mean and standard deviation of a weighted differences between each local feature and the $k$ Gaussian.
\end{itemize}

For those feature aggregation methods that require clustering, we project each local features into 256 dimension using PCA and the number of clusters for all encoding methods are 256, as suggested in \cite{xu2015discriminative}.

As can be seen in Table \ref{aggregate}, maximum pooling ($Max$) has similir or better performance compared to other methods. This observation is again different from \cite{lan2013cmu}, where the mean pooling ($Mean$) performs better than maximum pooling ($Max$). It is also interesting to find that $Mean\_std$ is consistently better than $Mean$. However, more complicated encoding methods such as $BoW$, $FV$ and $VLAD$ are all much worse than simple mean pooling. We conjecture that extracting more local features for each video and break each local features into lower dimension as in \cite{xu2015discriminative} will improve the results of these encoding methods. However,  it would incur excessive computational cost, which limits its usefulness in practice.

\subsection{How dense we need to do feature extraction?}

In studying the number of samples needed for each video, we also use maximum pooling ($Max$) as our feature aggregation method. The number of samples for each video ranges from 3 to 25. We also report the results of using maximum number of samples (Max), where we extract features for every frame/clip (for optical flow, we use a sliding window with step size equal to 1). On average,  there are 92 frames for each video in HMDB51 and 185 frames for each video in UCF101.

From the results shown in Table \ref{samples}, we can see that, after a certain threshold (15 in this case), the number of sampled frame/clip doesn't have great impact on the overall performance. A sample number of 25 is enough to achieve similar performance as densely sample every frame/clips. This is consistent with the observation in \cite{lan2013cmu} and largely because the information redundancy among frames.

\subsection{Comparison to the state-of-the-arts}

\begin{table}
\centering
\begin{tabular}{ |l | c |  c| }
 \hline
 & HMDB51 & UCF101 \\
 \hline
 IDT \cite{wang2013action} & 57.2 & 85.9 \\
 MIFS \cite{lan2014beyond} & 65.1 & 89.1 \\
 Two-stream \cite{simonyan2014two} & 59.4 & 88.0 \\
 TSN \cite{TSN2016} & 68.5& 94.0 \\
 Deep Quantization \cite{qiu2016deep} & - & 94.2\\
  Deep Quantization \cite{qiu2016deep} (w/ IDT) & - & 95.2\\
 TLE \cite{diba2016deep} &71.1 & 95.6\\
DOVF (ours) & 71.7&   94.9 \\
DOVF+ MIFS (ours) &  75.0 & 95.3 \\
 \hline
\end{tabular}
\caption{Comparison to the state-of-the-arts \label{state-of-the-art}}
\end{table}

In Table \ref{state-of-the-art}, we compare our best performance to the state-of-the-arts. Compared to the TSN \cite{TSN2016}, from which we improve upon, we get around 3\% and 1\% improvements on HMDB51 and UCF101 datasets, respectively. These results are much better than traditional IDT based methods \cite{lan2014beyond} and the original Two-stream CNNs \cite{simonyan2014two}. Compare to TLE \cite{diba2016deep} and Deep Quantization \cite{qiu2016deep}, our maximum pooling achieves similar results to their more complex bilinear models. We also show the results of fusing with MIFS \footnote{We download the the prediction scores of MIFS from \href{http://www.cs.cmu.edu/~lanzhzh/}{here}} using late fusion. For HMDB51, the improvement from fusing MIFS is very significant, we get more than 3\% improvement. The improvement for UCF101 is much smaller as its accuracy is in a place where is difficult to improve.

\section{Conclusions}

In this paper, we propose a method to get in-domain global video features by aggregating local neural network features. We study a set of problems including what features we should get, how to aggregate these local features into global features, and how dense we should extract these local features. After a set of experiments on UCF101 and HMDB51 datasets, we concludes that: 1) it is better to extract the outputs of the last convolution layer as  features; 2) maximum pooling generally work better than other feature aggregation methods including those that need further encoding; 3) a sparse sampling of more than 15 frames/clips per video is enough for maximum pooling. Although we present some observations about this new local features DOVF, the reasons behind these observations need further investigation. Also, the current two-stage approach only corrects the mistakes after it happens, we believe that a better way would be directly mapping a whole video into the video label, or so called end-to-end learning.  Our future works will focus on these two directions.

{\small
\bibliographystyle{ieee}
\bibliography{egbib_cvpr}
}

\end{document}